\documentclass[10pt,twocolumn,letterpaper]{article}

\usepackage{ijcb}
\usepackage{times}
\usepackage{epsfig}
\usepackage{graphicx}
\usepackage{amsmath}
\usepackage{amssymb}
\usepackage[norule,symbol,perpage]{footmisc}
\usepackage{caption}
\usepackage{subcaption}
\usepackage{tabularx}
\usepackage{threeparttable}
\usepackage{multirow}
\usepackage{comment}
\usepackage{makecell}
\usepackage{enumitem}
\usepackage{enumerate}
\usepackage{MnSymbol}
\usepackage{flushend}
\usepackage[pagebackref=true,breaklinks=true,colorlinks,bookmarks=false]{hyperref}

\ijcbfinalcopy % *** Uncomment this line for the final submission

\ifijcbfinal\fi

\makeatletter
\def\ps@IEEEtitlepagestyle{
\def\@oddfoot{\mycopyrightnotice}
\def\@evenfoot{}
}
\def\mycopyrightnotice{
{\hfill \footnotesize 978-1-7281-9186-7/20/\$31.00 \copyright 2020 IEEE\hfill}
}
\makeatother

\begin{document}

\title{Iris Liveness Detection Competition (LivDet-Iris) -- The 2020 Edition}

\author{ 
Priyanka Das\textsuperscript{$\dagger$}\textsuperscript{1},
Joseph McGrath\textsuperscript{$\dagger$}\textsuperscript{2},
Zhaoyuan Fang\textsuperscript{$\dagger$}\textsuperscript{2},
Aidan Boyd\textsuperscript{2}, 
Ganghee Jang\textsuperscript{1}, \\
Amir Mohammadi\textsuperscript{4},
Sandip Purnapatra\textsuperscript{1}, 
David Yambay\textsuperscript{1}, 
Sébastien Marcel\textsuperscript{4}, 
Mateusz Trokielewicz\textsuperscript{3},\\ 
Piotr Maciejewicz\textsuperscript{5}, 
Kevin Bowyer\textsuperscript{2}, 
Adam Czajka\textsuperscript{2}, 
Stephanie Schuckers\textsuperscript{1} \\
\textsuperscript{1}Clarkson University, 
\textsuperscript{2}University of Notre Dame, 
\textsuperscript{3}Warsaw University of Technology, Poland,\\ \textsuperscript{4}Idiap Research Institute, Switzerland, \textsuperscript{5}Medical University of Warsaw, Poland\\
{\small {\tt Organizers }\textsuperscript{$\dagger$}{\tt\small Equal Lead}}
\and
 Juan Tapia\textsuperscript{*}\textsuperscript{6},
 Sebastian Gonzalez\textsuperscript{*}\textsuperscript{9},
 Meiling Fang\textsuperscript{*}\textsuperscript{7,8}, 
 Naser Damer\textsuperscript{*}\textsuperscript{7,8}, 
 Fadi Boutros\textsuperscript{*}\textsuperscript{7,8}, 
 Arjan Kuijper\textsuperscript{*}\textsuperscript{7,8}\\
\textsuperscript{6}Universidad de Santiago - Chile , \textsuperscript{7}Fraunhofer Institute for Computer Graphics Research IGD, Darmstadt,\\ Germany, \textsuperscript{8}Department of Computer Science, Technical University of Darmstadt, Darmstadt,\\ Germany, \textsuperscript{9}TOC Biometrics - Chile\\
\textsuperscript{*}{\tt\small Competitors}
\and
Renu Sharma\textsuperscript{**}, Cunjian Chen\textsuperscript{**}, Arun Ross\textsuperscript{**}\\
Michigan State University\\
\textsuperscript{**}\tt\small Providers of the MSU baseline algorithms
}

\maketitle
\thispagestyle{empty}

\begin{abstract}
   Launched in 2013, LivDet-Iris is an international competition series open to academia and industry with the aim to assess and report advances in iris Presentation Attack Detection (PAD). This paper presents results from the fourth competition of the series: LivDet-Iris 2020. This year's competition introduced several novel elements: (a) incorporated new types of attacks (samples displayed on a screen, cadaver eyes and prosthetic eyes), (b) initiated LivDet-Iris as an on-going effort, with a testing protocol available now to everyone via the Biometrics Evaluation and Testing (BEAT)\footnote{\url{https://www.idiap.ch/software/beat/}} open-source platform to facilitate reproducibility and benchmarking of new algorithms continuously, and (c) performance comparison of the submitted entries with three baseline methods (offered by the University of Notre Dame and Michigan State University), and three open-source iris PAD methods available in the public domain. The best performing entry to the competition reported a weighted average APCER of 59.10\% and a BPCER of 0.46\% over all five attack types. This paper serves as the latest evaluation of iris PAD on a large spectrum of presentation attack instruments.

\end{abstract}

\let\thefootnote\relax\footnotetext{\mycopyrightnotice}

\section{Introduction}

Iris recognition systems have been deployed in 
commercial and government applications across the globe for more than two decades \cite{Jain16a}. Vulnerabilities of these systems against malicious attacks is an active area of research. One such attack that is being increasingly studied is the {\it presentation attack } (PA), where a sample is presented to the sensor with the goal of interfering with the correct operation of the system \cite{boyd2020iris}. Presentation attacks may be carried out with different motives: (1) to impersonate an identity during verification, (2) to conceal an identity during recognition or (3) to create a virtual identity during enrollment \cite{Hoffman2018}. Solutions to detect these attacks are referred to as {\it Presentation Attack Detection (PAD)} and include both hardware-based and software-based approaches. Software solutions are typically passive and mainly consider the static or dynamic features of the image or video presented to the system. Hardware solutions often employ active measurements of physical (color, density of tissue, optical properties) or physiological (pupil dilation) characteristics of the eye \cite{czajka2018presentation}. Research in presentation attack detection is an arms race: attacks on biometric systems are continually evolving, and system designers are continuously updating their security measures to efficiently detect artifacts as well as non-conformant uses of authentic biometric characteristics. {\em LivDet-Iris} is an international competition series launched in 2013 \cite{yambay2014schuckers}  to assess the current state of the art in iris PAD by the independent evaluation of algorithms and systems on data and artifacts not seen by the competitors when designing their solutions. This paper reports on the fourth edition of this competition: {\em LivDet-Iris 2020}. The most significant {\bf contributions of this paper} (and the LivDet-Iris 2020 competition itself) are:

\begin{itemize}[leftmargin=*,noitemsep]
\item A report on the current state-of-the-art in iris PAD based on independent testing of {\bf three algorithms submitted} to the competition organizers;
\item Introduction of three {\bf novel presentation attack instruments (PAI)}, when compared to previous LivDet editions: post-mortem iris images, electronic display, and fake/prosthetic/printed samples with add-ons. These attacks, combined with the printed iris images and the eyes with textured contact lenses, represent the {\bf five different PAIs} in the test set, \ie the largest spectrum of PAIs used to date in all iris PAD competitions.
\item {\bf Results from three different baseline methods} offered by the University of Notre Dame and the Michigan State University (see Sec.~\ref{Baseline} for details) and {\bf three open-source iris PAD methods} (see Sec. ~\ref{Opensource} for details).
\item Availability of the competition through the {\bf Biometrics Evaluation and Testing (BEAT)} \cite{anjos2017beat,marcel2013beat} platform (in addition to other algorithm submission options), implies some degree of privacy to the PAD algorithms as well as the test dataset.
\item {\bf Initiation of LivDet-Iris Competition as ``Ongoing''}, i.e., the competition benchmark will remain available to all researchers through the BEAT platform after this edition is concluded, which allows for testing all future algorithms according to the LivDet-Iris 2020 protocol, {\bf without revealing the test data}. 
\end{itemize}

\section{Performance Evaluation Metrics}
\label{evaluation}

LivDet-Iris 2020 follows the recommendations of ISO/IEC 30107-3 \cite{ISO_IEC_301073:2017} in employing two basic PAD metrics in its evaluations:

\begin{itemize}[leftmargin=*, noitemsep]
\item \textbf{Attack Presentation Classification Error Rate (APCER)}, the proportion of attack presentations of the same PAI species incorrectly classified as bonafide presentation, \ie spoof classified as live, and
\item \textbf{Bonafide Presentation Classification Error Rate (BPCER)}, the proportion of bonafide presentations classified as attack presentations, \ie live classified as spoof.
\end{itemize}

Both the APCER and BPCER metrics are used to evaluate the algorithms. ISO also recommends to use the maximum value of APCER when multiple PA species (or categories) are present in case of system-level evaluation, which is primarily designed for industry applications. This, however, is inconsistent with our prior competitions \cite{yambay2014schuckers,7947701,yambay2017livdet} and also our goal to consider the detection of all PAIs, and not to rank the competitors by looking at their worst-performing PA. Thus, we introduced the weighted average of APCER over all PAIs:

\begin{itemize}[leftmargin=*, noitemsep]
\item \textbf{Weighted Average of APCER} (APCER$_{\mbox{\footnotesize average}}$), which is the average of APCER accross all PAIs, weighted by the sample counts in each PAI category, as reported in Table~\ref{table:Dataset}.

\noindent
Only for the {\bf purpose of competition ranking}, the Average Classification Error Rate (ACER) was computed to select the winner:  

\item \textbf{Average Classification Error Rate (ACER):} the average of APCER$_{\mbox{\footnotesize average}}$ and BPCER.
\end{itemize}

Note that ACER has been deprecated in ISO/IEC 30107-3:2017 \cite{ISO_IEC_301073:2017} in the industry-related PAD evaluations.

\section{Iris PAD Evaluation Efforts To Date}

Iris PAD literature offers a wide spectrum of software- and hardware-based solutions, and two recent survey papers \cite{boyd2020iris,czajka2018presentation} provide a comprehensive overview of the current state of the art. In this section, we offer a summary of all known public iris PAD evaluation efforts to date.

\subsection{MobILive}
{\it Mobile Iris Liveness Detection Competition (MobILive)} was held in 2014 to assess the state of art of algorithms for iris liveness detection for {\bf mobile applications}. The competition concentrated on the simplest PAI: printed iris images of an authorized subject presented to the sensor by an unauthorized subject. The purpose of the competition was to assess the performance of algorithms to distinguish between live iris images and paper iris printouts. The best performing algorithm achieved the mean of APCER and BPCER equal to 0.25\% \cite{sequeira2014mobilive}.

\subsection{LivDet-Iris 2013, 2015 and 2017}

LivDet-Iris 2013 \cite{yambay2014schuckers} was the first public evaluation platform for advancements in iris PAD focused on systems and algorithms employing {\bf ISO-compliant iris images} (in particular, acquired in near infrared light). The competition subsequently occurred in 2015 \cite{7947701} and 2017 \cite{yambay2017livdet}. Every LivDet-Iris competition offered both software- and system-level evaluation. The software-based competition evaluates the performance of {\em algorithms} in the task of detection of presentation attacks. The system-based competition evaluates the performance of {\em complete systems} (including sensors) against physical presentation attacks. Over the past editions, ten participants submitted algorithms and none elected to submit for a system-level evaluation. Table~\ref{table:LivDet_Series} summarizes all editions, including the current (2020) installment.

\begin{table*}[!ht]
\footnotesize
\centering
\caption{LivDet-Iris Competition Series Summary}
\label{table:LivDet_Series}
\begin{tabular}{|l|l|c|c|c|c|}
\hline
\textbf{Competition year} & \textbf{Presentation Attack Instruments}  & \textbf{New train / test data} & \textbf{Number of} & \multicolumn{2}{c|}{\textbf{Best performance}} \\
& {\bf in test data} & {\bf delivered by organizers} & {\bf competitors} & APCER & BPCER \\\hline\hline
2013 & Printed Irises, Patterned Contact Lenses & Yes / Yes &  3 & 5.7\% & 28.6\% \\
\hline
2015 & Printed Irises, Patterned Contact Lenses & Yes / Yes & 4 & 5.48\% & 1.68\% \\
\hline 
2017 & Printed Irises, Patterned Contact Lenses & Yes / Yes & 3 & 14.71\% & 3.36\% \\
\hline\hline
2020 & Printed Irises, Patterned Contact Lenses, & & & &  \\
(reported in this paper) & Fake/Prosthetic/Printed Eyes with Add-ons & No / Yes  & 3 & 59.10\% & 0.46\%\\
& Eyes Displayed on Kindle, Cadaver Irises & & & & \\
\hline
\end{tabular}
\end{table*}

\subsection{LivDet-Iris 2020}
The LivDet-Iris 2020 competition was launched in May 2020 and was co-organized by five organizations: Clarkson University (USA), University of Notre Dame (USA), Warsaw University of Technology (Poland), IDIAP Research Institute (Switzerland) and Medical University of Warsaw (Poland). In previous editions, this competition had two parts: \textit{Algorithms} and \textit{Systems}. The \textit{Algorithms} part required participants to submit their algorithm(s) to the organizers for independent testing on the unknown test data. The \textit{Systems} section required submission of a complete system, including hardware, designed for presentation attack detection. After submission, our team evaluated the submitted systems based on varied physical attack types. One winner was selected for each part of the competition based on the average performance of detecting spoofs and accepting live samples. Participation was encouraged from all academic and industrial institutions. In contrast to past LivDet-Iris competitions, the 2020 edition did not offer any official training data -- the competitors were free to use any proprietary and/or publicly available data to design their algorithms. In this 2020 edition, for the first time, an open-source research experimentation platform, BEAT was used to host the competition. The BEAT platform facilitates further evaluations of algorithms by any researcher, using the identical test data and protocol as in this competition edition, even after the completion of the competition.

It is important to note that the entire LivDet-Iris competition series focuses on evaluation of capabilities of algorithms to {\bf generalize to  unknown circumstances}. While a very brief characterization of attack types included in the test set is provided to the competitors, the test samples are not revealed, and therefore not used for training. All evaluations are completed by the LivDet-Iris organizers, and are {\it not} self-reported by participants.

\section{Experimental Protocol and Evaluation}
\label{Protocol}

\subsection{Participation} \label{participation}
Participation in LivDet-Iris 2020 was open to all academic and industrial institutions with the option to participate anonymously in both the \textit{Algorithms} and \textit{Systems} part. Anonymous participation allowed the competitors to retain their identity from the co-authors' list. Fourteen teams registered for the competition from across the globe. The organizers received three algorithm submissions from three registered teams. There were no submissions to the systems portion of the competition. All three competing teams were invited to contribute to this report by describing their PAD methods briefly.
\subsection{Datasets}\label{Dataset}

\paragraph{Training dataset} LivDet-Iris 2020 was different from previous editions in that the organizers did not announce any official training set. Instead, the participants were encouraged to use all data available to them (both publicly available and proprietary) to make their solutions as effective and robust as possible. The entire past LivDet-Iris benchmarks were also made publicly available \cite{yambay2014schuckers,7947701,yambay2017livdet}. Additionally, the competition organizers shared 5 examples of each PAI (and these samples were not used later in evaluations) to familiarize the competitors with the test data format (pixel resolution, bits per pixel used to code the intensity, etc.).
\graphicspath{{images/}}
\begin{figure*}[!ht]
\centering

\subcaptionbox{\centering Paper printout}{\includegraphics[width=0.24\textwidth]{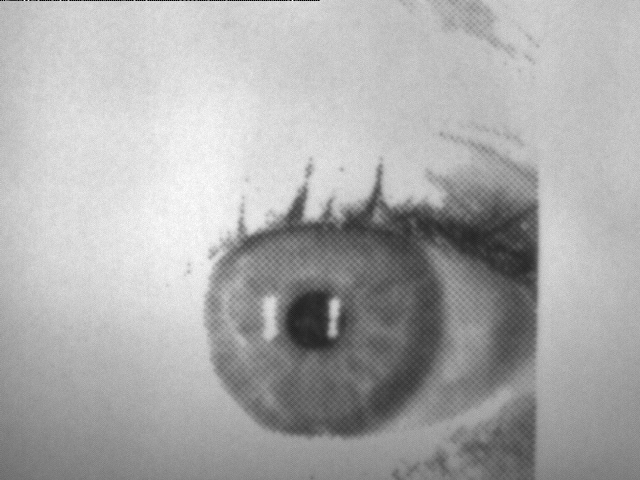}}
\hfill
\subcaptionbox{\centering Cosmetic contact lens \newline on the live eye}{\includegraphics[width=0.24\textwidth]{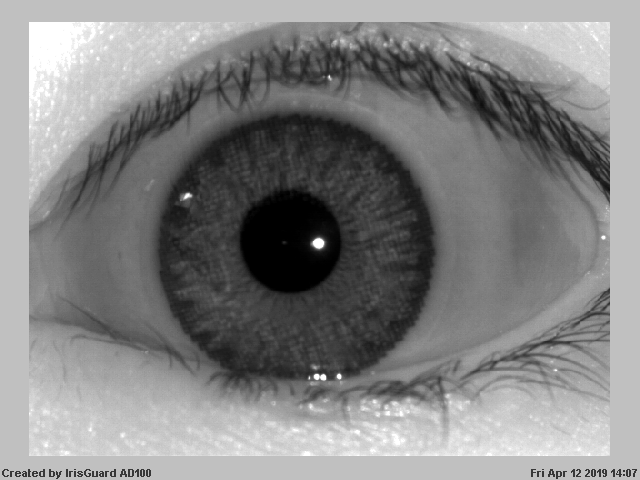}}
\hfill 
\subcaptionbox{\centering Cosmetic contact lens \newline on the printed eye}{\includegraphics[width=0.24\textwidth]{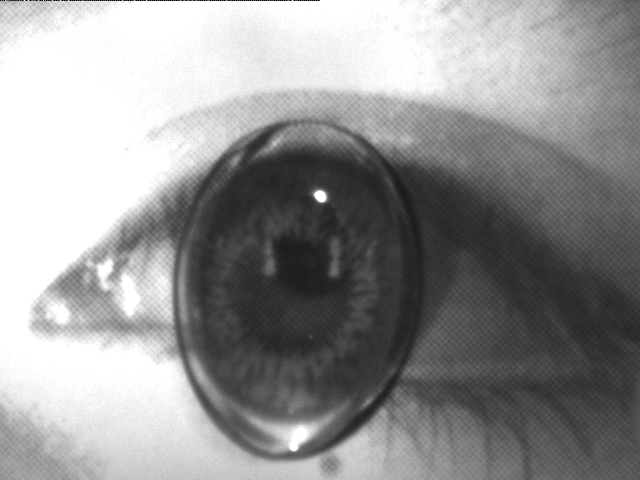}}
\hfill
\subcaptionbox{\centering Eye dome on the printed eye}{\includegraphics[width=0.24\textwidth]{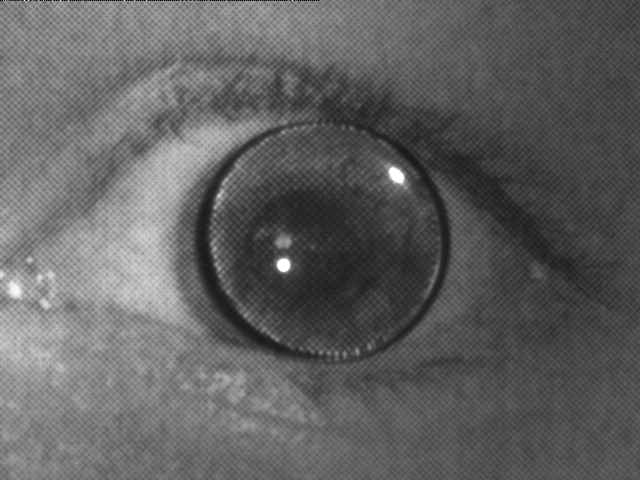}}
\hfill
\subcaptionbox{\centering Electronic display}{\includegraphics[width=0.24\textwidth]{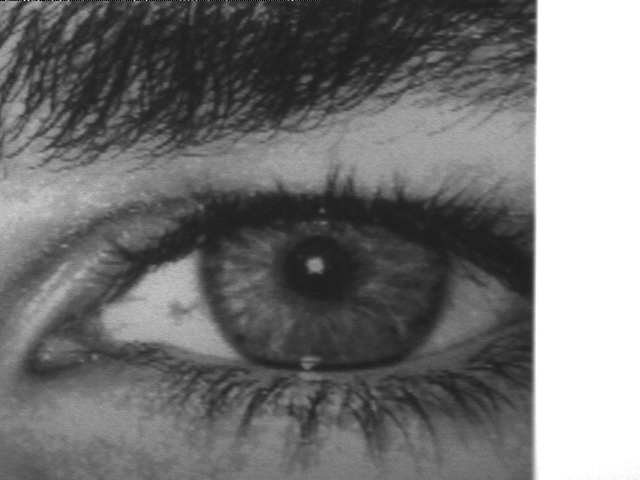}}
\hfill
\subcaptionbox{\centering Doll eye}{\includegraphics[width=0.24\textwidth]{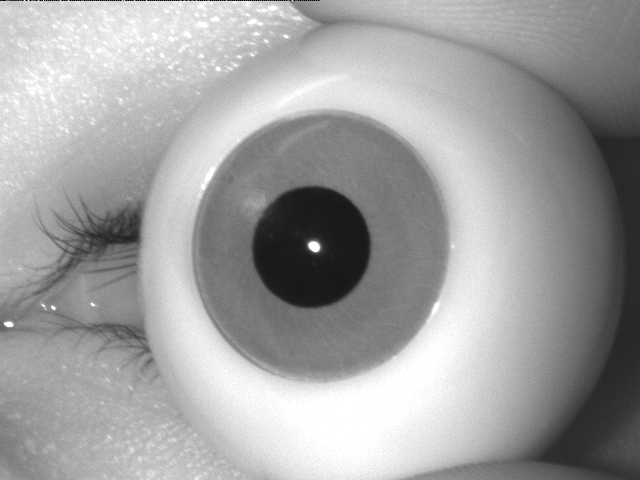}}
\hfill
\subcaptionbox{\centering Cosmetic contact lens \newline on the doll eye}{\includegraphics[width=0.24\textwidth]{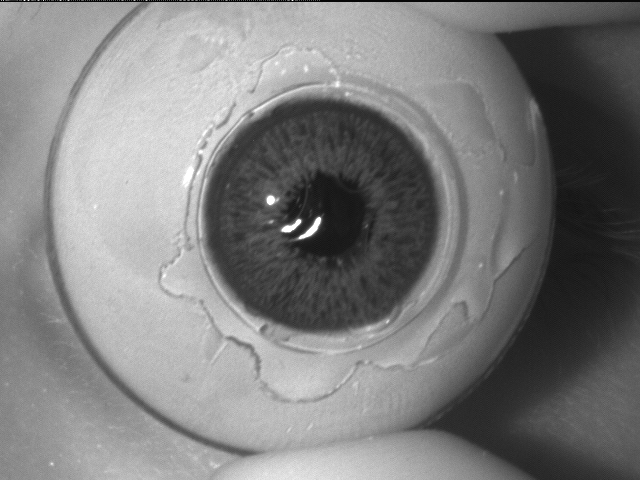}}
\hfill
\subcaptionbox{\centering Cadaver eye}{\includegraphics[width=0.24\textwidth]{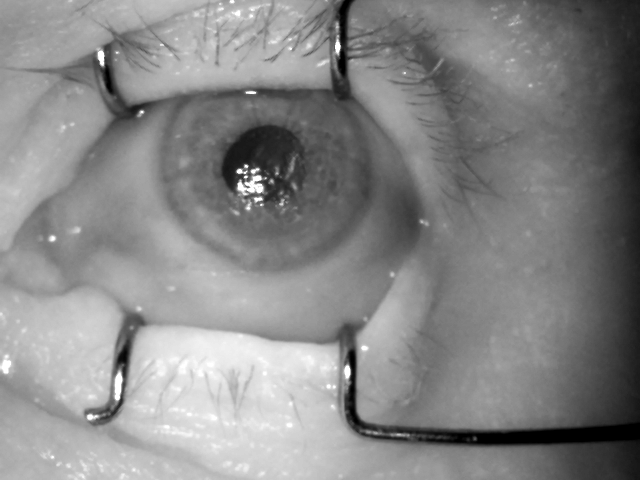}}
\caption{Example images of all presentation attack types present in the LivDet-Iris 2020 test dataset.}
\label{fig:Example_images}
\end{figure*}

\paragraph{Test dataset} The testing set employed in this competition was a combination of data from all three organizers: Clarkson University (CU), University of Notre Dame (ND) and Warsaw University of Technology (WUT). The dataset consisted of 12,432 images (5,331 live and 7,101 spoof samples), as summarized in Table~\ref{table:Dataset}. Sample images from the dataset are shown in Figure~\ref{fig:Example_images}. Five Presentation Attack Instruments (PAI) categories were included in the dataset:

\begin{itemize}[leftmargin=*, noitemsep]
\item \textbf{Printed eyes:} 1,049 samples created using five different printers (Epson Stylus Pro 9900, HP M652, Xerox C60, OKI MB-471, Cannon Super G3) and two different print qualities (``office'' and ``professional''). Two different paper types were used (matte and glossy paper). Images were collected with the Iris ID iCAM7000.% were used for the printed eye attack types. 
\item \textbf{Textured contact lenses:} 4,336 samples were acquired using LG IrisAccess 4000 and IrisGuard AD100 under different illumination setups offered by these sensors (two different illuminants in LG 4000 and six different illuminants in AD 100). This portion of the data were collected from 88 subjects (176 irises) wearing cosmetic contact lenses of three different brands: Johnson \& Johnson, Ciba Vision, and Bausch \& Lomb. 

\item \textbf{Eyes displayed on Kindle e-Ink:} 81 such samples were captured in NIR spectrum the Iris ID iCAM7000 sensor.    

\item \textbf{Fake/Prosthetic/Printed Eyes with Add-ons:} This category has five sub-categories of spoofs making up a total of 541 samples, captured in NIR spectrum by the Iris ID iCAM7000 sensor, with non-uniform distribution of samples in each category:
 \begin{itemize}[leftmargin=*, noitemsep]
     \item \textbf{Textured Contacts on Printed Eyes:} patterned contact lenses added on top of the printed eye images.
     \item \textbf{Textured Contacts on Doll Eyes:} patterned contact lens put on the iris area of plastic doll eyes.
     \item \textbf{Clear Contacts on Printed Eyes:} transparent contact lens added on top of the printed eye images.
     \item \textbf{Eye Dome on Printed Eyes:} transparent 3D plastic eye domes added on top of the printed eye images. 
     \item \textbf{Doll Eyes:} Fake eyes of two different types -- Van Dyke Eyes (has higher iris quality details) and Scary eyes (plastic fake eyes with simple pattern on iris region); different color variation of both types of fake eyes were included.
 \end{itemize}
 \item \textbf{Cadaver Eyes:} The Warsaw-BioBase-Post-Mortem-Iris v3.0 dataset \cite{TROKIELEWICZIMAVIS2019, WarsawColdIris3} encompasses a total of 1,094 NIR images (collected with an IriShield M2120U handheld iris recognition camera) and 785 visible light images (obtained with Olympus TG-3) collected from 42 post-mortem subjects, and is fully subject-disjoint from previous Warsaw post-mortem dataset publicly available before LivDet-Iris 2020 competition \cite{WarsawColdIris1, WarsawColdIris2}. Data collection sessions were organized accordingly with medical staff and cadaver availability and ranges from several hours after demise up to 369 hours postmortem. For the purpose of LivDet-Iris 2020 competition, only NIR images are employed. This data collection had institutional review board clearance and the ethical principles of the Helsinki Declaration were followed by the data acquisition staff. 
\end{itemize}

\begin{table*}[!ht]
\footnotesize
\centering
\caption{Test Dataset Summary}
\label{table:Dataset}
\begin{tabular}{|c|c|c|c|}
\hline
\textbf{Class} & \textbf{Presentation Attack Instruments}  & \textbf{Sample Count} & \textbf{Sensor}  \\
\hline
Live & - & 5,331 &  LG 4000, AD 100, Iris ID iCAM7000  \\
\hline
Spoof & Printed Eyes & 1,049 & Iris ID iCAM7000 \\
\hline 
Spoof & Textured Contact Lens & 4,336 & LG 4000, AD 100,
Iris ID iCAM7000 \\
\hline
Spoof & Electronic Display & 81 & Iris ID iCAM7000 \\
\hline 
Spoof & Fake/Prosthetic/Printed Eyes with Add-ons & 541 & Iris ID iCAM7000 \\
\hline
Spoof & Cadaver Iris &1,094 &	IriTech IriShield\\
\hline

\end{tabular}
\end{table*}

\subsection{Experimentation Platform}\label{BEAT}
LivDet-Iris 2020 used the benefits of the Biometrics Evaluation and Testing (BEAT) platform for the competition. BEAT is a solution for open access, scientific information sharing and re-use including data and source code while protecting privacy and confidentiality. It allows easy online access to experimentation and testing in computational science. The platform allows access and comparison of different experimentation and results. \cite{anjos2017beat,marcel2013beat}. Participants were encouraged to submit their algorithm through this platform. However, we also accepted the submissions that were sent to us for evaluation as executables for Windows, Linux, or Mac OS. Alternatively, we accepted also codes in Python (v.3.7) or MATLAB (2019a or above). All submission options were equivalent in terms of participation. There are several advantages of conducting this edition of LivDet-Iris in the BEAT platform:

\begin{itemize}[leftmargin=*, noitemsep]
    \item {\bf Privacy:} The algorithms submitted by the participants remain invisible to everybody except the participant. Similarly, any data uploaded to the BEAT platform also remain inaccessible to any user of the platform. In particular, this allowed us to share the test data in an anonymous and reproducible manner that otherwise could not be shared, due to sponsor restrictions. 
     \item {\bf Re-submission:} The participants can make multiple submissions before the deadline but do not have access to results until after their final submission.
    \item {\bf Continuity:} This platform will serve as an iris PAD ``on-going'' benchmark after LivDet-Iris 2020 is concluded, since the test data and protocol are planned to be retained on BEAT and available for executing algorithms by all interested researchers.
\end{itemize}

\subsection{LivDet Iris 2020 Competition Algorithms}\label{LivDet_Algo}
All teams were given the opportunity to submit a description of their submitted algorithm, and two such descriptions are provided below. One team elected not to provide a description.

\textbf{USACH/TOC Team:} For this competition an algorithm was presented based on a multilabel CNN network that has been used to detect printed images and patterned contact lenses. The SMobileNet and FMobilNet models are both based on MobilenetV2. SMobileNet was trained from scratch to detect the presence of patterned contact lenses in the iris image area. FMobilNet was trained using fine-tuning with average and max pooling options, in order to detect the printed images of the whole image by identifying the physical source of the image. Finally, a multi-output classifier was developed in order to identify fake or live or real images. This option allowed the team to create a lightweight classifier to be implemented in a mobile iris recognition camera such as Gemini Iritech.

\textbf{FraunhoferIGD Team:} The algorithm starts by finding special local-features in the investigated image. These local-features are clustered into a number of classes. Moreover, an image patch is extracted from the area around each of these local-features. For each of the clusters, a classifying network is used to determine the origin of the patch (belonging to this specific cluster) as a bonafide image or an attack image. After that, a logistic regression model was trained for each cluster class. This logistic regression takes the classification probability from the network and the cluster class reliability and results in a final bonafide/attack classification score. All the classification scores produced by the logistic regressions from different patches are fused using a simple mean-rule. The algorithm uses the K-means approach to build the local-feature clusters and calculate the class reliability. The used patch size is $64\times64$ pixels. The classification network used is trained from scratch and is based on the MobileNetV3-Samll \cite{Howard_2019_ICCV} neural network architecture.

\subsection{Baseline Algorithms} \label{Baseline}
All organizing teams who contributed to the baseline and open-source algorithm performance evaluation had access to LivDet Iris 2020 test dataset. However, as a declaration, every team that contributed a baseline/open source algorithm verified that they did not use the data from the test dataset as part of training.

\textbf{Notre Dame PAD Algorithm:} The implemented solution extends the methodology proposed by Doyle and Bowyer \cite{Bowyer_COMPUTER_2014} and the feature extraction is based on Binary Statistical Image Features (BSIF) proposed by Kannala and Rahtu \cite{Kannala_ICPR_2012}. In this method, the calculated ``BSIF code'' is based on filtering the image with $n$ filters of size $s\times s$, and then binarizing the filtering results with a threshold at zero. Hence, for each pixel $n$ binary responses are given, which are in the next step translated into a $n$-bit grayscale value. The histograms resulting from gray-scale BSIF codes are later normalized to a $z$-score and used as texture descriptors with the number of histogram bins equal to $2^n$. We use a \textit{Best Guess} segmentation technique to select a region of interest. A separate set of three classifiers (SVM, RF, and MLP) was trained on the Notre Dame LivDet-Iris 2017 dataset for each feature set ($n,s$ pair). Since not all the classifiers have the same strength, a subset of the strongest classifiers was selected through testing on the Clarkson LivDet-Iris 2017 dataset and majority voting is applied to these selected classifiers to come up with a final decision.

\textbf{MSU PAD Algorithm 1:} The proposed algorithm, namely TL-PAD~\cite{ChenR18,btasChenR18}, operates on the cropped iris regions and offers a simple and fast solution. It utilizes the pre-trained ImageNet model to initialize the weights and then performs transfer learning. First, an off-line trained iris detector~\cite{ChenR18} was used to obtain a rectangular region encompassing the outer boundary of the iris. Then, the iris region was automatically cropped based on the estimated rectangular coordinates. Finally, the cropped iris region was input to a CNN to train the iris PA detection model. MobileNetV2 was used as the backbone network with squeeze-and-excitation module applied to the last convolution layer to recalibrate channel-wise features. The training was fine-tuned on an existing ImageNet model, by leveraging extensive data augmentation schemes, including rotation, shift and flip operations, to name a few. The learning rate was set to 0.0001 and Adam optimizer was used. The algorithm was trained on a proprietary dataset comprising of 12,058 live images and 10,622 PA images.

\textbf{MSU PAD Algorithm 2:} MSU second baseline method is a variant of D-NetPAD \cite{Sharma2020} whose base architecture is Dense Convolutional Network 161 (DenseNet161) \cite{Huang2017}. The input to the model is a cropped iris region resized to 224 $\times$ 224. The model weights are first initialized by training on the ImageNet dataset \cite{Deng2009} and then fine-tuned using bonafide iris and PA samples. Fine-tuning was performed with a proprietary dataset, NDCLD-2015 \cite{NDCLD2015} and Warsaw PostMortem v3 dataset. The proprietary dataset consists of 19,453 bonafide irides and 4,047 PA samples. PA samples include 51 kindle display attacks, 1,005 printed eyes, 1,804 artificial eyes, and 1,187 cosmetic contact lenses. From the NDCLD-2015 dataset, 2,236 cosmetic contact lenses images were used for the training, and from the Warsaw PostMortem v3 dataset, 1,200 cadaver iris images from the first 37 cadavers were used. The architecture consists of 161 convolutional layers integrated into four Dense blocks, and three Transition Layers lie between the Dense blocks. The last layer is a fully connected layer. A detailed description of the architecture is provided in \cite{Huang2017}. The learning rate used for the training is 0.005, the batch size is 20, the number of epochs is 50, the optimization algorithm is stochastic gradient descent with a momentum of 0.9, and the loss function used is cross-entropy.

\subsection{Open Source Algorithms}\label{Opensource}

For completeness, three open-source iris PAD algorithms available today in the public domain (in addition to the baseline algorithms) are also evaluated. All three algorithms are trained on a subset of the 2017 LivDet-Iris competition data, constructed such that 100 samples are taken from each unique combination of data provider, image label, and dataset partition (\eg one possible combination would be Notre Dame, contact lens, and the training set).

\textbf{RegionalPAD:} Hu \etal \cite{hu2016iris} investigate the use of regional features in iris PAD. Features are extracted from local neighborhoods based on spatial pyramid (multi-level resolution) and relational measures (convolution on features with variable-size kernels). Several feature extractors such as Local Binary Patterns (LBP), Local Phase Quantization (LPQ), and intensity correlogram are investigated. In our experiments, we use the three-scale LBP-based feature, since it achieves the best performance as pointed out by the original authors.

\textbf{SIDPAD:} Gragnaniello \etal \cite{gragnaniello2016using} proposes that the sclera region also contains important information about iris liveness. Hence, the authors extract features from both the iris and sclera regions. The two regions are first segmented and scale-invariant local descriptors (SID) are applied. A bag-of-feature method is then used to summarize the features. A linear Support Vector Machine (SVM) is used to perform final prediction. We refer to this method as SIDPAD.

\textbf{DACNN:} Gragnaniello \etal \cite{gragnaniello2016biometric} incorporates domain-specific knowledge of iris PAD into the design of their model. With the domain knowledge, a compact network architecture is obtained and regularization terms are added to the loss function to enforce high-pass / low-pass behavior. The authors show that the method can detect both face and iris spoofing attacks. We refer to this method as DACNN.

\section{Results and Analysis}\label{results}

\begin{figure*}[!ht]
\centering

\subcaptionbox{\centering LivDet-Iris 2020 competitors}{\includegraphics[width=0.32\textwidth]{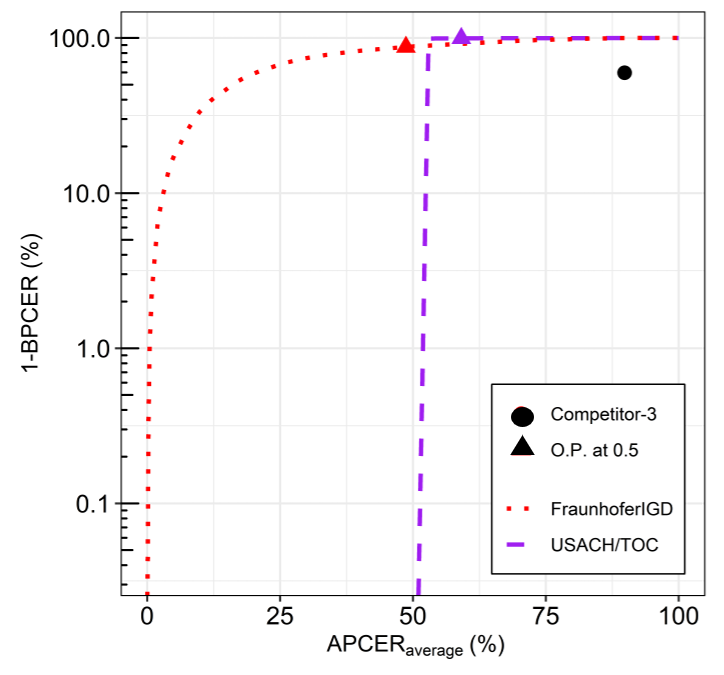}}
\hfill
\subcaptionbox{\centering Open-source algorithms} {\includegraphics[width=0.32\textwidth]{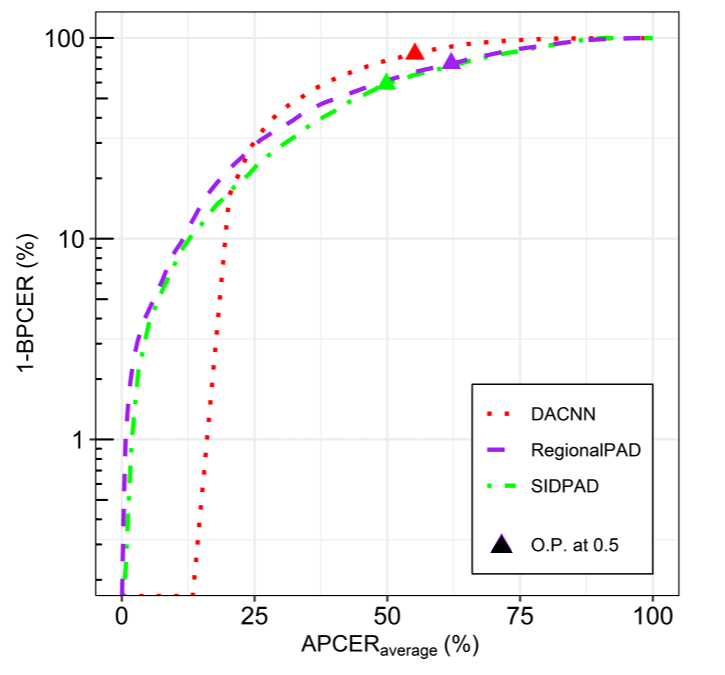}}
\hfill 
\subcaptionbox{\centering Baseline algorithms}{\includegraphics[width=0.31\textwidth]{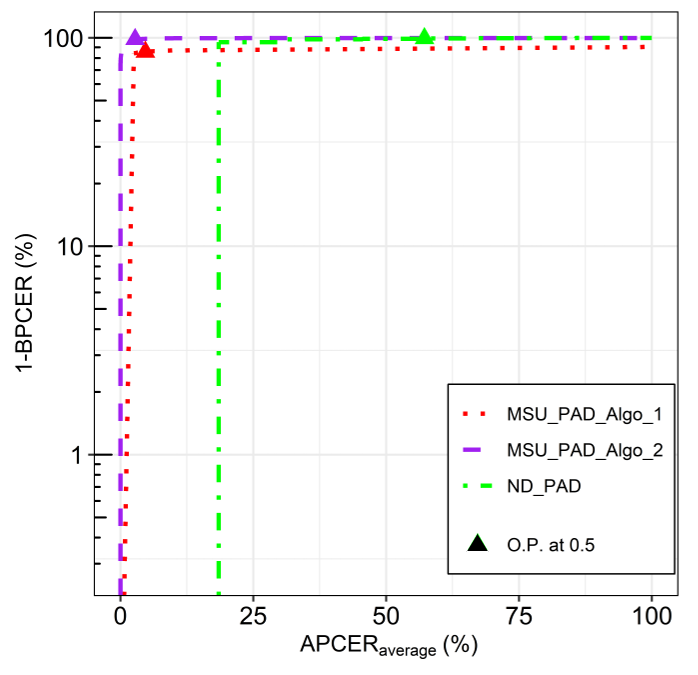}}

\caption{ROC curves for all nine algorithms presenting the overall performance on samples representing all five PAIs. The overall APCER is evaluated based on (APCER$_{\mbox{\footnotesize average}}$). The operating point (``O.P. at 0.5'') used to rank participants of this LivDet-Iris competition is marked by a $\filledmedtriangleup$ on each curve.}
\label{fig:ROC_category}
\end{figure*}

\begin{figure*}[!ht]
\centering

\subcaptionbox{\centering Printed Eyes}{\includegraphics[height= 3.1cm]{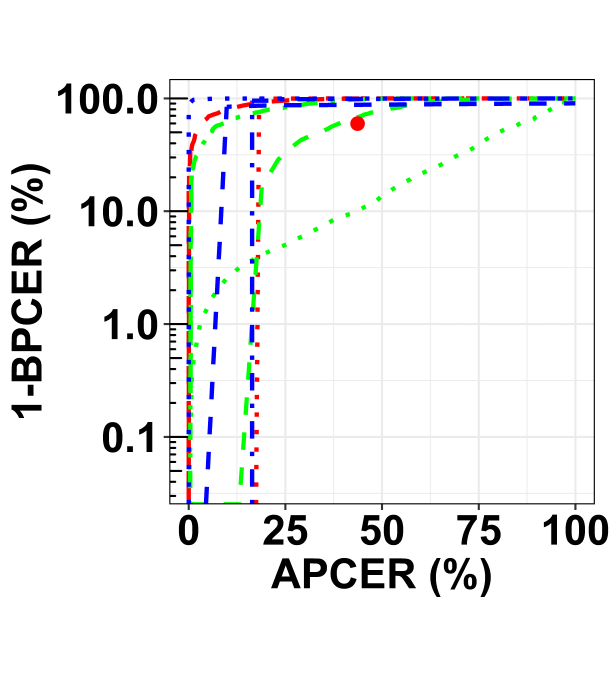}}
\hfill
\subcaptionbox{\centering Contact Lenses} {\includegraphics[height= 3.1cm]{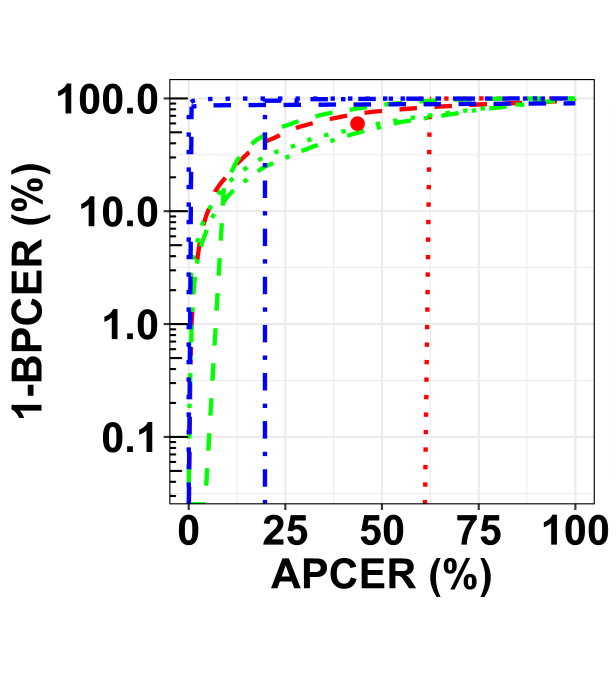}}
\hfill 
\subcaptionbox{\centering Fake/Prosthtic Eyes with Add-ons}{\includegraphics[height= 3.1cm]{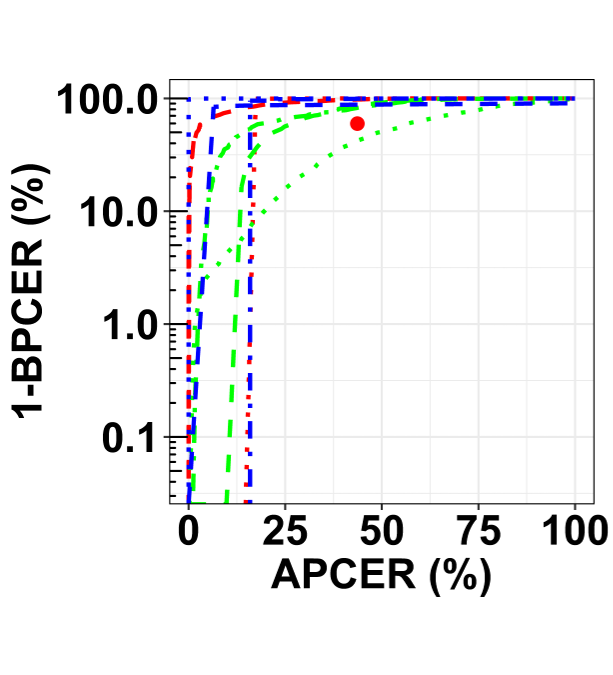}}
\hfill
\subcaptionbox{\centering Electronic Display}{\includegraphics[height= 3.1cm]{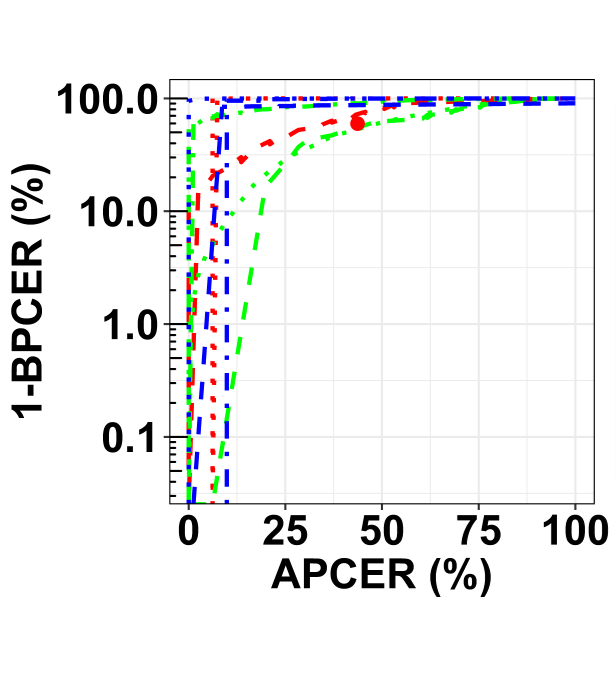}}
\hfill
\subcaptionbox{\centering Cadaver Irises}{\includegraphics[height= 3.1cm]{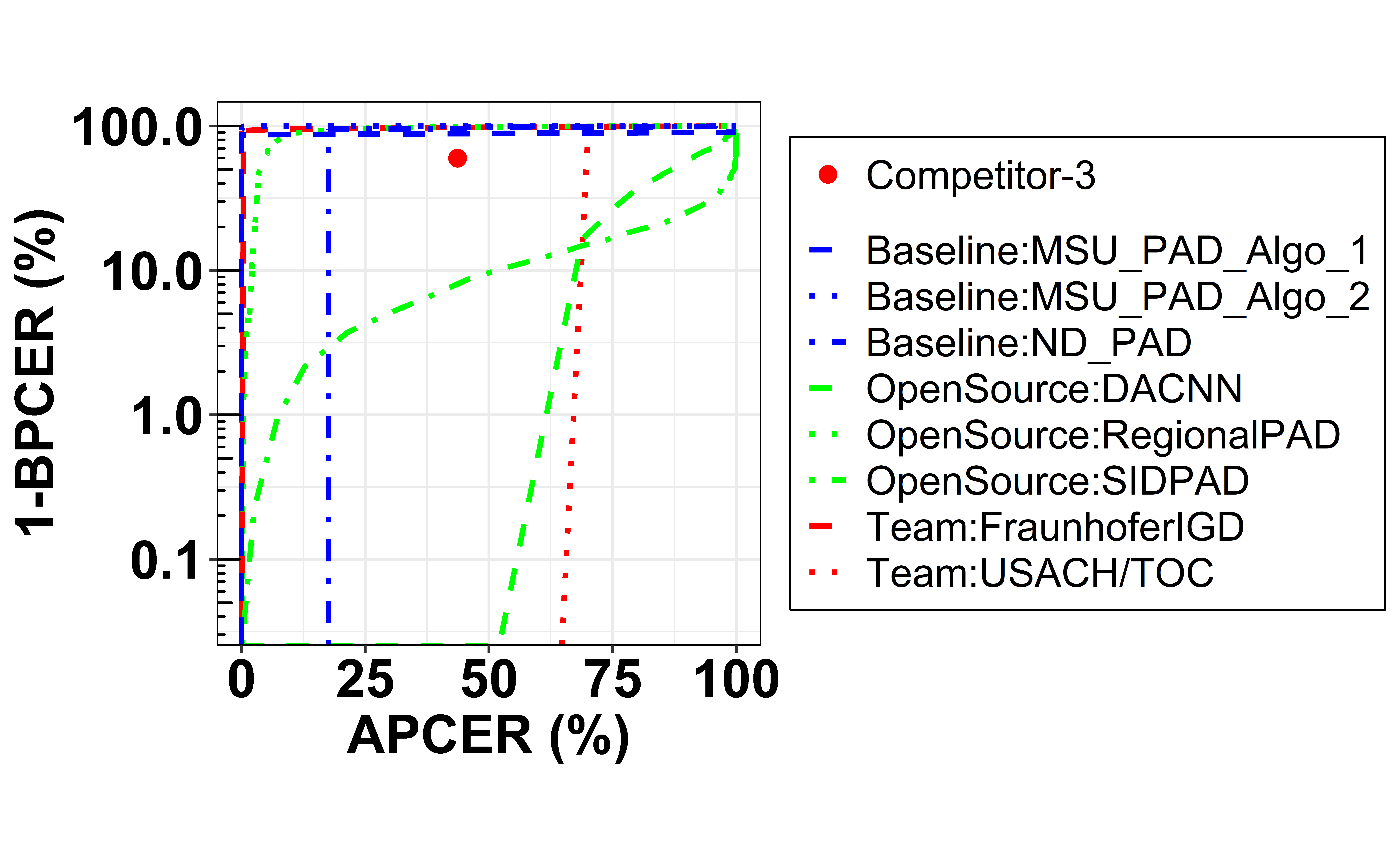}}
\caption{Same as in Fig. \ref{fig:ROC_category}, except that the performance of all nine methods is presented separately for each PAI.}
\label{fig:ROC_PAI}
\end{figure*}

This section discusses the performance of the algorithms in 3 categories: (1) LivDet-Iris 2020 competitors, (2) baseline algorithms, and (3) open-source algorithms. The performance has been evaluated based on APCER for each of the five PAIs. APCER and BPCER are evaluated at the threshold of 0.5, which was announced prior to the competition. A summary of the error rates for all 3 categories is provided in Table~\ref{table:Performance}. The ROCs shown for all PAIs broken by algorithm category (competitors, open-source and baselines) are shown in Figure~\ref{fig:ROC_category}. The ROCs for individual PAIs for all nine methods are depicted in Figure~\ref{fig:ROC_PAI}. Below we discuss the performance of algorithms in three groups of methods.

 \begin{table*}[!t]
 \centering
 \begin{threeparttable}
 \footnotesize
\centering
\caption{Error Rates (\%) for all algorithms calculated at a threshold of 0.5, corresponding to each PAI (ACPER) and the overall performance (ACER)}
\label{table:Performance}
\begin{tabular}{|*{10}{c|}}
\hline
\textbf{Method} & \multirow{2}{*}{\textbf{Algorithm}} & \multicolumn{5}{c|}{\textbf{APCER}}  & \multicolumn{2}{c|}{ \textbf{Overall Performance}} & \multirow{2}{*} {\textbf{ACER}} \\
\cline{3-9}
\textbf{category} &  & \textbf{PE} & \textbf{CL} & \textbf{ED} & \textbf{F/P} & \textbf{CI} & \textbf{APCER$_{\mbox{\footnotesize average}}$} & \textbf{BPCER} & \\
 \hline\hline
 \multirow{3}{*}{\makecell{\textbf{Livet Iris 2020}\\{ \textbf{Submissions}}}} &  Team: USACH/TOC & 23.64 & 66.01 & 9.87  & 25.69 & 86.10 & 59.10 & 0.46 & 29.78 \\
 \cline{2-10}
   &Team: FraunhoferIGD & 14.87 & 72.80 & 53.08 &  19.04 & 0 & 48.68 & 11.59 & 30.14 \\
 
   \cline{2-10}
  & Competitor-3 & 72.64 & 43.68 & 83.95 & 73.19 & 89.85 & 57.8  & 40.31 & 49.06 \\
    \hline
    \hline
   \multirow{3}{*}{\textbf{Baselines\textsuperscript{*}}} & ND PAD\textsuperscript{**} & 55.95 & 50.74
 & 35.80 & 43.25 & 92.59 & 57.21 & 0.71 & 28.96  \\
 \cline{2-10}
   & MSU PAD Algorithm 1 & 14.96 &2.23 & 23.45 & 10.90& 0 & 4.67  & 0.56 & 2.61 \\
 \cline{2-10}
   & MSU PAD Algorithm 2 & 2.38 & 3.85 & 1.23 & 0.18 & 0.18  & 2.76  & 1.61 & 2.18 \\
 \hline
\hline
\multirow{3}{*}{\textbf{Open Source}}  &  DACNN\textsuperscript{**} & 54.53 & 45.94 & 75.31 & 41.22 & 97.99 & 55.2 & 16.39
  & 35.8 \\ 
\cline{2-10}
 & SIDPAD\textsuperscript{**}  & 8.48 & 52.19 & 1.24 & 17.93 & 99.82 & 49.85 & 39.96 & 44.9\\
\cline{2-10}
& RegionalPAD\textsuperscript{**} & 92.18 & 67.62 & 96.29 & 70.79 & 6.49
& 62.42 & 23.80 & 43.11 \\
\hline
\end{tabular}
\begin{tablenotes}
      \small
      \item \textsuperscript{*}Authors had access to the entire LivDet-Iris 2020 test dataset, but did not use it as part of training. \textsuperscript{**}These methods were {\bf not} trained on all categories of PAIs present in the LivDet-Iris 2020 test dataset.  \textbf{PE:} Printed Eyes; \textbf{CL:} Textured Contact Lens; \textbf{ED:} Electronic Display; \textbf{F/P:} Fake/Prosthetic/Printed Eyes with Add-ons; \textbf{CI:} Cadaver Iris.
    \end{tablenotes}
    \end{threeparttable}
\end{table*}
\textbf{LivDet Iris 2020 Competitors:} %Three teams submitted their PAD algorithms for the competition- USACH/TOC, FraunhoferIGD and Competitor -3. 
Team USACH/TOC was determined as the winner based on lowest ACER = 29.78\%, very closely followed by Team FraunhoferIGD with ACER = 30.14\%, and Competitor-3 with an ACER of 49.06\%. The winning team's method achieved also the lowest BPCER = 0.46\% out of all nine algorithms in the three categories. This aligns well with the operational goal of PAD algorithms to correctly detect bonafide presentations (\ie, and not to contribute to system's False Rejection Rate) and capture as many attacks as possible. The three algorithms had variable performance for each type of PAI. The algorithm offered by USACH/TOC was specifically tuned for printed eyes (PE) and trained from scratch to detect textured contact lenses (CL) (as explained in Sec.~\ref{LivDet_Algo}), but performed best for the electronic display (ED) PAI achieving APCER=9.87\%, which is lower than all competing algorithms (53.08\% and 83.95\%) by a large margin. Algorithm offered by Fraunhofer IGD, performed best in three categories: printed eyes (APCER = 14.87\%), fake/prosthetic eyes (APCER = 19.04\%) and cadaver irises (perfect detection of all cadaver samples). However, the detection of bonafide samples was worse (BPCER = 11.59\%) compared to the winner (BPCER=0.46\%). The algorithm offered by Competitor-3 was the lowest for the contact lenses (CL) category with APCER = 43.68\%. It is important to note that all of these results are based on independent evaluation, the competitors did not have an access to test data, and trained their algorithms on data not necessarily representing all PAIs present in the test data. It demonstrates the difficulty of open-set iris PAD.\\

\textbf{Baseline Algorithms:} The results of baseline algorithms are included to additionally demonstrate how a good representation of PAIs during training is important. The baseline offered by the University of Notre Dame was trained solely on live samples and contact lens PAI. It is thus not surprising to see an overall performance (across all PAIs) to be close to the competition winner. The weak performance on CL PAI category also suggests lower generalization capabilities of this image texture-based method onto unknown contact lens brands and patterns. In contrast, two baselines developed by Michigan State University offer the best performance out of all 9 algorithms. This could be due to the use of a more comprehensive training set to design the methods. In particular, MSU PAD Algorithm 2 resulted in a weighted APCER of 2.76\% at a BPCER of 1.61\%. 

\textbf{Open Source Algorithms:} All three algorithms lacked balance in the performance between bonafide and attack samples, and achieved high BPCER ranging between approximately 16\% and 40\%. The SIDPAD algorithm, which considers both the sclera and iris portions of the eye in their algorithm design to detect presentation attacks, performed well for three PAIs in comparison to other algorithms: printed eyes (PE) with APCER of 8.48\%, electronic display (ED) with an APCER of 1.24\% and fake/prosthetic/printed eyes with add-ons with APCER of 17.93\%. However, SIDPAD demonstrated limited accuracy of bonafide detections (approx. 40\% of BPCER) and failed in recognizing cadaver irises. The RegionalPAD algorithm, which considered three-scale LBP based features, achieved a low error rate only with cadaver iris (CI) attack type with APCER of 6.49\%, while it failed to detect reliably printed eyes and electronic display attacks. The DACNN algorithm, even though it demonstrated a relatively low ACER in the open-source category, it presented limited capability of detecting all PAIs, and also relatively high BPCER = 16.39\%. This may suggest that some of the older iris PAD methods, available in the public domain, may offer lower accuracy when applied to currently observed attacks.

\section{Conclusions}
The LivDet-Iris 2020 featured multiple new additions to the evaluation of iris presentation attack detection: (a) employed three novel PAIs (cadaver iris, fake/prosthetic/printed eyes with various add-ons, and electronic display), (b) introduced BEAT platform for the competition thereby facilitating privacy, re-submission and continuity of algorithm evaluation in the public domain, and (c) provided a comparative analysis of the nine state-of-art methods (three competition algorithms, three baseline algorithms and three open-source algorithms). The winning entry performed with an ACER of 29.67\% (BPCER = 0.46\ and APCER averaged over all PAIs = 59.10\%). However, two baseline algorithms from MSU resulted in the best performance.

We note a degradation in the best overall performance of the winning entry in LivDet-Iris 2020 compared to the previous competitions organized in 2013, 2015 and 2017 (as shown in Table~\ref{table:LivDet_Series}). This degradation can be attributed to multiple factors:

\begin{itemize}[leftmargin=*, noitemsep]
    \item[a)] highly increased complexity in the test dataset: five different PAI categories were employed in the competition this year compared to two PAIs in previous editions;
    \item[b)] introduction of novel attack types with limited or no access to large-enough public datasets for a few PAIs; 
    \item[c)] no specific training dataset was offered, and that design choice was left to be decided by competitors;
    \item[d)] the results could reflect variability between the training and the test datasets in terms of environmental factors, sensors, quality of PAIs, and the use of ``unseen" PAIs. 
\end{itemize}

The results from this competition indicate that iris PAD is still far from a fully solved research problem. Large differences in accuracy among baseline algorithms, which were trained with significantly different data, stress the importance of access to large and diversified training datasets, encompassing a large number of PAIs. We believe that this competition, and the benchmark now available to researchers via the BEAT platform, will contribute to our efforts as a biometric community in winning the PAD arms race. 

\section{Acknowledgement}
{\small This material is based upon work supported in part by the National Science Foundation under Grant No. \#1650503 and the Center for Identification Technology Research. 
MSU's and CU's research is supported in part by the Office of the Director of National Intelligence (ODNI), Intelligence Advanced Research Projects Activity (IARPA), via IARPA R\&D Contract No. 2017 - 17020200004. The views and conclusions contained herein are those of the authors and should not be interpreted as necessarily representing the official policies, either expressed or implied, of ODNI, IARPA, or the U.S. Government. The U.S. Government is authorized to reproduce and distribute reprints for governmental purposes notwithstanding any copyright annotation therein.}

{\small
\bibliographystyle{ieee}
\bibliography{bibliography.bib}
}

\end{document}